\documentclass[11pt]{article}
\usepackage[margin=1in]{geometry}
\usepackage{amsmath, amssymb, amsthm}
\usepackage{natbib}
\usepackage{hyperref}
\usepackage{authblk}
\usepackage{graphicx} 
\usepackage{booktabs}
\usepackage[english]{babel}
\usepackage[autostyle, english = american]{csquotes}
\MakeOuterQuote{"}

\title{Active Inference with Reusable State-Dependent Value Profiles}
\author{Jacob Poschl}
\date{November 2025}

\begin{document}

\maketitle
\section{Abstract}
Adaptive behavior in volatile environments requires agents to deploy different value-control regimes across latent contexts, but representing separate preferences, policy biases, and action confidence for every situation is intractable. We introduce value profiles: a small set of reusable bundles of value-related parameters—outcome preferences, policy priors, and policy precision—that are assigned to hidden states in the generative model. As posterior beliefs over states evolve trial-by-trial, effective control parameters emerge through belief-weighted mixing, enabling state-conditional strategy recruitment without maintaining independent parameters for each situation. We evaluate this framework in probabilistic reversal learning, comparing static precision, entropy-coupled dynamic precision, and profile-based models using cross-validated log-likelihood and information criteria. Model comparison using AIC favors the profile-based model over simpler alternatives ($\approx$100-point differences), with consistent parameter recovery demonstrating structural identifiability even when context must be inferred from noisy observations. Model-based inference suggests that, in this task, adaptive control operates primarily through policy prior modulation rather than policy precision modulation, with gradual belief-driven profile recruitment confirming state-conditional rather than merely uncertainty-driven control. Overall, reusable value profiles provide a tractable computational account of belief-conditioned value control in volatile environments, providing a reusable, mode-like representational scheme for behavioral flexibility that yields testable signatures of belief-conditioned control.
\section{Introduction}
How should agents adapt \emph{value control} when environmental statistics change? In volatile environments, agents benefit from control regimes that support rapid strategy revision and information seeking; in stable environments, the same regime can yield over-sensitivity to noise and unnecessary exploration. Predictive processing and hierarchical Bayesian accounts formalize this trade-off as inference under a generative model, where \emph{precision} (i.e., confidence-weighting of prediction errors) governs the gain on prediction errors and thereby shapes learning, attention, and action selection \cite{parr2022,mathys-2011,Schwartenbeck2016,feldman-2010}. However, while many models posit that agents adjust control-relevant precisions (e.g., policy precision $\gamma$) in response to inferred context (e.g., volatility), the computational principles that determine \emph{which control parameters should change, and how they should be assigned across latent contexts}, remain underspecified. Moreover, in active inference implementations, key control parameters are often treated as global or context-level quantities, which can be powerful but empirically difficult to identify and difficult to map to mechanistic constraints.

This paper introduces \emph{value profiles}: a state-conditional mechanism for reusable, identifiable control that does not require learning separate parameters for every situation. Rather than tuning control parameters globally or independently for each context, agents maintain a small library of profiles---reusable bundles of outcome preferences ($C$), policy priors ($E$), and policy precision ($\gamma$), allowing multiple control channels to be recruited jointly, with task demands and identifiability determining which channels are expressed in fitted behavior. These parameters constitute distinct computational channels for modulating effective learning and action (e.g., by shaping preferences over outcomes and biases over policies, and by sharpening or flattening policy selection). Profiles are assigned to hidden states in the generative model, and as posterior beliefs over states evolve trial-by-trial, effective parameters emerge through belief-weighted mixing. This belief-conditioned, state-conditional recruitment of bundled control settings enables adaptive behavior in changing environments while remaining structurally identifiable from behavior.

The profile framework is theoretically motivated by a fundamental constraint: agents cannot maintain independent preferences and precisions for every situation they encounter. Instead, they must compress the value space into a manageable set of reusable modes. Profiles provide this compression by bundling parameters into coherent behavioral strategies that can be shared across multiple hidden states. A small number of profiles can support flexible behavior across a much larger state space through belief-weighted recruitment.

In tasks requiring rapid belief revision, such as probabilistic reversal learning, agents with profile-based value control should exhibit context-dependent behavioral strategies that emerge from belief-weighted profile recruitment. The specific adaptive mechanisms—whether through policy precision ($\gamma$), outcome preferences (C), or policy priors (E)—remain an empirical question. These adaptations should be belief-weighted and direction-specific, reflecting which state the agent infers rather than merely how uncertain it is, producing behavioral signatures that are difficult to capture with global or entropy-coupled mechanisms.

Our contributions are threefold. First, we introduce \emph{value profiles} as a parameter-sharing construction for state-conditional control: reusable bundles of control-relevant parameters assigned to latent context states and recruited through belief-weighted mixing. Second, we derive how profile recruitment induces smoothly varying effective preferences and priors over policies (and, when needed, effective policy precision). Third, in a probabilistic reversal-learning task with hidden context inferred from reward statistics, we test whether this structure yields identifiable and predictively useful behavioral signatures; cross-validated model comparison favors profile-based control relative to static and entropy-coupled precision baselines. Finally, model-based inference indicates that, in this task, adaptation is expressed primarily through policy preferences/priors (e.g., hint-seeking) rather than requiring context-specific shifts in policy precision $\gamma$.

\section{Background}
\subsection{Predictive Processing and the Free Energy Principle}
Predictive processing formalizes perception, action, and learning as inference under a generative model \cite{friston2009, friston2010, parr2022}. The brain predicts sensory causes and updates beliefs when predictions fail, with precision weighting determining which prediction errors drive learning and how decisively policies are selected \cite{Schwartenbeck2016}. The Free Energy Principle provides the normative foundation: agents minimize prediction error (or equivalently, variational free energy) through both perceptual inference and active sampling  \cite{friston2010}. However, computational principles for assigning precision remain underspecified \cite{Vance2017, Sajid2021}. Rather than treating precision as a global parameter, we propose that agents maintain reusable value profiles—bundles of preferences and precisions assigned to hidden states and mixed according to current beliefs.

\subsection{Belief and Value Structures}
Agents construct beliefs about hidden states through inference over sensory observations, updating their generative models as evidence accumulates \cite{parr2022}. However, beliefs alone are insufficient for adaptive behavior—an agent must also represent which states are desirable, which outcomes warrant attention, and how decisively to act \cite{Friston2015-gp}. Values provide this evaluative structure, determining what matters and how strongly it matters within the current context. This aligns with goal-directed formulations of cognition, where categorization and decision-making are conceived as means to achieve preferred goal states rather than ends in themselves \cite{rigoli2017}.

The environment presents a fundamental capacity problem: the space of possible states and contingencies far exceeds what any bounded agent could represent with independent parameters. If each situation required its own preference structure and precision setting, the resulting parameter space would be intractable for neural storage and for learning from finite experience. Because learning and planning with large/unknown state spaces already strain parameter capacity \cite{doshi-velez2009}, we extend the same parsimony to value parameters by sharing them via reusable profiles rather than per-state tuning.

Reusable value profiles offer a compressive solution. Rather than independent parameterizations, agents maintain a small repertoire of coherent behavioral modes (e.g., exploration, exploitation, threat response) that can be deployed across multiple contexts. Each mode bundles together preferences (which outcomes to pursue), policy biases (which actions to favor), and precision (how decisively to act), because these parameters naturally co-vary when implementing unified behavioral strategies \cite{Friston2015-gp}. An exploratory mode, for instance, couples low decisiveness with preferences for informative outcomes and biases toward novel actions; an exploitative mode couples high decisiveness with outcome maximization and familiar action selection.

This bundling is not arbitrary but reflects the structure of adaptive behavior in which attention, evaluation, and action selection must be coordinated to implement meaningful strategies. By reusing a finite set of profiles across a larger state space, agents achieve computational efficiency while maintaining behavioral flexibility. This approach extends the factorization rationale of goal-directed Bayesian models, in which context, category, and action representations are distinct yet hierarchically related, and where model evidence trades off accuracy and complexity \cite{rigoli2017}.

Precision in this paper refers to the broader predictive-processing notion of gain control—any mechanism that modulates the influence of prediction errors \cite{Kanai2015-vh, feldman-2010}. This includes outcome preferences (C), action biases (E), and policy confidence. Standard discrete active inference often isolates policy precision as $\gamma$, an inverse-temperature over expected free energy. In our profile framework, $\gamma$ may differ across profiles in principle, but our results show that context-dependent behavior is recoverable even when fitted profiles converge to similar $\gamma$, indicating that policy preferences/priors can be the dominant adaptive channel in this task.

\subsection{Active Inference Formalization}
We implement this theoretical framework within discrete active inference, where agents maintain generative models over hidden states and observations \cite{Smith2022-eo}. For our purposes, we distinguish between belief parameters that define the agent's world model (likelihood mappings A, transition dynamics B, state priors D) and value parameters that determine preferences and action selection (outcome preferences C, policy priors E, and policy precision $\gamma$). While this distinction is somewhat artificial—outcome preferences C are often treated as priors within the generative model—it proves useful for isolating which aspects of the model adapt across behavioral modes.

Value profiles bundle these preference and precision parameters (C, E, $\gamma$) into coherent units that can be assigned to hidden states, while the world model (A, B, D) remains shared across profiles. As beliefs about states fluctuate, effective values emerge through weighted mixtures of the assigned profiles. This provides a tractable way to implement state-conditional value control: rather than tuning preferences and precision independently for each state, agents reuse a small set of behavioral modes that share a common world model but differ in their evaluative and control parameters.

The specific assignment of active inference parameters to "belief" versus "value" categories is a modeling choice open to alternative interpretations. In hierarchical active inference architectures, for instance, value structures could themselves be represented across multiple levels, with higher-level context states modulating lower-level preferences and precisions. Similarly, the decision to bundle C, E, and $\gamma$ together reflects our hypothesis about which parameters co-vary during behavioral mode switching, but other groupings are theoretically possible. For this work, we adopt a single-level (non-hierarchical) active inference model as a proof of concept, demonstrating that state-conditional profile mixing can improve behavioral coherence in volatile environments. Extensions to hierarchical models with nested value representations remain an important direction for future work.

\subsection{Related Work and Biological Grounding}
Neural control appears to rely on coordinated, context-sensitive modulation across multiple circuits rather than uniform global adjustment \cite{Kanai2015-vh, Ferguson2020-fs}. Neuromodulatory systems---including dopaminergic projections from ventral tegmental area and substantia nigra, cholinergic inputs from basal forebrain, and norepinephrine from locus coeruleus---have been implicated in regulating the \emph{gain} on sensory prediction errors as well as the \emph{selectivity} and \emph{decisiveness} of action selection (e.g., dopaminergic modulation of prediction-error weighting in hierarchical active inference models; \cite{Friston2012-PLOS}, average-reward/opportunity-cost accounts of tonic dopamine and response vigor \cite{niv-2007}, and adaptive gain modulation in the LC--NE system that trades off exploitation and exploration; \cite{Aston-Jones-2005}). Evidence suggests these systems operate in a context-sensitive manner: dopamine signals show phasic responses to unexpected outcomes that can promote behavioral adjustment, while tonic levels correlate with exploitation and goal-directed persistence, and track average reward rate to modulate response vigor in humans \cite{Beierholm-2013}. Thalamic nuclei (i.e., pulvinar and mediodorsal) can gate cortical information flow by modulating the gain on ascending prediction errors \cite{Sherman-2003, Kanai2015-vh, feldman-2010}. In primates, single-unit recordings show that dopamine neurons initially respond to unexpected rewards, then transfer their phasic response to reward-predicting cues, and exhibit well-timed pauses when expected rewards are omitted---a pattern predicted by temporal-difference reward prediction error models \cite{Schultz-1997}. Critically, these mechanisms do not appear to reduce to a single global parameter: they exhibit state-dependent and task-dependent modulation, suggesting the brain maintains multiple operating regimes that can be flexibly recruited \cite{Friston2012-PLOS, Schultz-1997}, in line with Bayesian accounts that link acetylcholine and norepinephrine to expected and unexpected uncertainty \cite{yu-2005}, and hierarchical Bayesian models in which inferred volatility controls precision-weighted learning \cite{mathys-2011}. Complementing neuromodulatory accounts, recordings in primate prefrontal cortex show that recurrent dynamics can select which of several co-present inputs (e.g., motion vs.\ color) are integrated depending on task context, despite weak early gating of irrelevant inputs \cite{Mante2013}.

Active inference provides a formal framework for \emph{adaptive control} through probabilistic inference, in which multiple parameter families can modulate effective learning and action. In particular, gain on sensory prediction errors shapes belief updating, while parameters governing policy selection shape how deterministically actions are chosen based on expected outcomes. Contemporary implementations have explored several approaches: dynamic gain/precision adjustment based on inferred volatility or belief uncertainty, hierarchical schemes in which different levels maintain partially independent gain parameters, and learned schedules that adapt through experience. These models capture phenomena such as the explore--exploit trade-off and attentional modulation. However, many implementations either (i) treat key control parameters as global quantities shared across contexts or (ii) couple adaptation to generic uncertainty measures such as entropy over beliefs. While computationally tractable, these approaches face two limitations in complex environments with multiple distinct contexts: they can require many degrees of freedom when extended to richly structured tasks, and they do not naturally capture the discrete, recognizable behavioral modes---such as threat response, exploratory foraging, or habitual control---observed in biological systems and human behavior.

Computationally, assigning independent gain or control parameters to each state/context can create identifiability problems: with limited behavioral data, a large number of weakly constrained parameters cannot be reliably estimated. Conceptually, purely global or uncertainty-coupled mechanisms are also insufficient to explain rapid switching between qualitatively distinct strategies when latent states change (e.g., shifting from cautious exploration to confident exploitation following a single informative observation). We posit that biological systems may address this through reusable, coordinated configurations rather than continuously retuning a single gain parameter: the brain may maintain a repertoire of neuromodulatory states that jointly shape attention, valuation, and action selection into coherent modes, echoing adaptive gain accounts of LC--NE phasic and tonic modes linked to exploitation and exploration \cite{Aston-Jones-2005}.

Our contribution addresses this gap by formalizing \emph{value profiles} within active inference: small sets of reusable parameter bundles (outcome preferences, policy biases, and \emph{optionally} policy precision) that are assigned to hidden states and mixed according to current beliefs. This provides a state-conditional mechanism for \emph{belief-conditioned value control} that is learnable from behavioral data and computationally efficient through parameter sharing, while remaining broadly consistent with the idea that neuromodulatory systems coordinate distinct behavioral regimes. The framework yields testable predictions in behavior: as inferred context shifts, the agent should exhibit belief-weighted transitions in effective control parameters (e.g., in policy biases and information-seeking tendencies), producing mode-like changes that distinguish profile recruitment from global or purely uncertainty-coupled alternatives and thereby help bridge computational models and neurobiological observations.

Lastly, the value profile framework relates to hierarchical active inference architectures but operates at a single level of representation. Where hierarchical models achieve context-dependent control through explicit multi-level inference with higher-level states modulating lower-level preferences, profiles achieve similar effects by identifying reusable patterns of value parameters and recruiting them based on beliefs within a single (potentially multi-factor) state space. This provides computational efficiency: rather than maintaining and updating beliefs at multiple hierarchical levels, agents perform standard inference over states, then use those beliefs to mix a small repertoire of behavioral modes. The tradeoff is flexibility---hierarchical models can represent nested temporal abstractions and compositional goal structures that profiles cannot directly capture in their current form---against tractability---profiles require fewer parameters and less computational overhead than full hierarchical inference. This positions the framework as an intermediate solution: more adaptive than static global control, more tractable than full hierarchical architectures.

\section{Methodology}
\subsection{Generative Model}
We model behavior within a discrete partially observable Markov decision process (POMDP) framework. At each trial $t$, the agent observes outcomes $o_t \in \{1,\ldots,O\}$, infers hidden states $s_t \in \{1,\ldots,S\}$, and selects actions $a_t \in \{1,\ldots,A\}$. The generative model is specified by three standard components: the likelihood matrix $\mathbf{A} \in [0,1]^{O \times S}$ encoding $p(o|s)$, the transition matrix $\mathbf{B} \in [0,1]^{S \times S \times A}$ encoding $p(s'|s,a)$, and the prior $\mathbf{D} \in [0,1]^S$ encoding $p(s_1)$.\footnote{For multi-factor state spaces and multi-modality observations, $\mathbf{A}$, $\mathbf{B}$, and $\mathbf{D}$ generalize to object arrays with factor-specific or modality-specific components. We present single-factor notation for clarity; our implementation uses the full multi-factor formulation.} These belief parameters define the agent's world model. In standard active inference, value parameters include outcome preference logits $\mathbf{C} \in \mathbb{R}^O$ (unnormalized log-priors over outcomes), policy prior logits $\mathbf{E} \in \mathbb{R}^\pi$ (unnormalized log-biases over action sequences, where $\pi$ denotes the number of policies), and policy precision $\gamma \in \mathbb{R}_{>0}$ (inverse temperature for action selection). Our contribution modifies how these value parameters are structured and deployed.

\subsection{Value Profiles}
Rather than maintaining global or context-wide value parameters, we propose that agents maintain $K$ distinct value profiles. Each profile $k \in \{1,\ldots,K\}$ bundles three components:
\begin{equation}
\Omega_k = \{\mathbf{C}_k, \mathbf{E}_k, \gamma_k\}
\end{equation}
where $\mathbf{C}_k \in \mathbb{R}^O$ are outcome preference logits, $\mathbf{E}_k \in \mathbb{R}^\pi$ are policy prior logits, and $\gamma_k \in \mathbb{R}_{>0}$ is policy precision. Outcome preferences determine which observations are desirable, policy priors encode habitual action biases, and policy precision controls how deterministically the agent selects actions. These parameters are bundled because they co-vary when implementing coherent behavioral strategies: an exploratory profile might combine low precision (stochastic action selection) with preferences for informative outcomes and biases toward novel actions, while an exploitative profile couples high precision with outcome maximization and familiar action selection. The logits $\mathbf{C}_k$ and $\mathbf{E}_k$ are mean-centered to fix the softmax gauge and improve identifiability, then converted to probability distributions via softmax normalization when used during inference.

\subsection{State Assignment and Belief-Weighted Mixing}
Profiles are assigned to hidden states through an assignment matrix $\mathbf{Z} \in [0,1]^{S \times K}$, where each row sums to one: $\sum_{k=1}^K Z_{s,k} = 1$ for all $s$. Hard assignment sets each row to be one-hot (each state uses exactly one profile), while soft assignment allows convex mixing. 

For multi-factor state spaces, profile assignment targets a specific state factor that represents behaviorally relevant context. In our implementation, profiles are assigned to the context factor (volatile vs. stable), while other factors (arm identity, action state) share profiles based on context beliefs. At each trial, the agent's posterior beliefs over the assignment-relevant state factor $q_t(s_{\text{ctx}})$ determine which profiles are recruited. Profile weights are computed by pooling beliefs through the assignment matrix:
\begin{equation}
w_t(k) = \sum_{s_{\text{ctx}}=1}^{S_{\text{ctx}}} q_t(s_{\text{ctx}}) Z_{s_{\text{ctx}},k}
\end{equation}
These weights form a probability distribution over profiles: $w_t \in [0,1]^K$ with $\sum_k w_t(k) = 1$. Effective trial-wise parameters emerge through belief-weighted mixing:
\begin{equation}
\mathbf{C}_t^{\text{eff}} = \sum_{k=1}^K w_t(k) \mathbf{C}_k, \quad \mathbf{E}_t^{\text{eff}} = \sum_{k=1}^K w_t(k) \mathbf{E}_k, \quad \gamma_t^{\text{eff}} = \sum_{k=1}^K w_t(k) \gamma_k
\end{equation}
When beliefs strongly favor a single context state, the weighting concentrates on that state's assigned profile. When context beliefs are uncertain, multiple profiles contribute proportionally, producing intermediate effective parameters. This mechanism makes value control state-conditional while maintaining continuous adaptation through belief-weighted mixing. We mix categorical natural parameters ($\mathbf{C}_k$ and $\mathbf{E}_k$ logits) linearly, yielding smooth interpolation between profile-defined action-value landscapes under uncertain context beliefs; for $\gamma_k$, linear mixing defines an effective inverse temperature under the same uncertainty.

\subsection{Inference and Control}
State inference is performed via standard categorical filtering under the assumed generative model. Given the previous posterior $q_{t-1}(s)$ and action $a_{t-1}$, a predictive prior is formed by propagating beliefs through the transition matrix: $\tilde{q}_t(s) = \sum_{s'} \mathbf{B}[a_{t-1}]_{s|s'} q_{t-1}(s')$. Upon observing $o_t$, the posterior is updated as $q_t(s) \propto \mathbf{A}[o_t|s] \tilde{q}_t(s)$ with normalization. Throughout, $\mathbf{A}$, $\mathbf{B}$, and $\mathbf{D}$ are treated as fixed (task-known) belief parameters, while value-profile parameters are fit from behavior. This posterior is then marginalized over the assignment-relevant factor and mapped through $\mathbf{Z}$ to compute profile weights $w_t$, which yield effective parameters via mixing.

\begin{equation}
p(\pi \mid \text{history}_t) \propto 
\exp\left(-\gamma_t^{\text{eff}}\, G_t(\pi) + E_{t,\pi}^{\text{eff}}\right),
\end{equation}
where $E_{t,\pi}^{\text{eff}}$ denotes the effective \emph{log prior} (logit) for policy $\pi$ and $G_t(\pi)$ denotes its expected free energy.\footnote{We use the standard active-inference definition of expected free energy, in which $G_t(\pi)$ decomposes into (i) expected divergence between predicted outcomes and preferences (risk) and (ii) an information-gain term (epistemic value). Action selection samples a policy from this posterior and executes its first action.}

\subsection{Model Specification, Optimization, and Identifiability}
The full model is parameterized by $\Theta = \{\mathbf{C}_k, \mathbf{E}_k, \gamma_k, \mathbf{Z}\}_{k=1}^K$, where each profile $k$ has its own outcome preference logits, policy prior logits, and policy precision, along with the assignment matrix linking profiles to states. In the current work, we adopt an estimation strategy optimized for model recovery and parameter identifiability. For the generative phase, profile parameters are specified a priori based on task structure (volatile vs.\ stable contexts), allowing us to generate data with known ground-truth parameter values. For the recovery phase, we employ exhaustive grid search over discretized parameter spaces (detailed in Section 5.3.3) rather than continuous optimization. This two-stage coarse-to-fine grid search systematically evaluates the parameter space, is robust to local minima, and enables direct assessment of whether behavioral data can recover the generative parameter structure. While computationally expensive, this approach prioritizes reliability of parameter recovery over computational efficiency.

In principle, profile parameters could also be estimated via MAP optimization of $\mathcal{L}(\Theta) = \sum_t \log p(a_t \mid \Theta) + \log p(\Theta)$ using constrained gradient-based methods; we leave this as future work. 

To ensure identifiability, we impose structural constraints that remove degeneracies in the parameter space. Preference and policy prior logits are mean-centered before softmax transformation, eliminating translation invariance. Policy precisions are constrained to be positive ($\gamma_k > 0$) through the search space definition. Assignment matrix rows are constrained to the probability simplex ($\sum_k Z_{s,k} = 1$). These constraints remove common degeneracies (e.g., translation invariance of logits) and reduce non-identifiability by restricting parameters to a canonical gauge, thereby limiting transformations that can yield indistinguishable behavior.

As in other mixture-like constructions, the model is invariant to permuting profile labels (label switching): relabeling $k$ leaves the likelihood unchanged unless an ordering or anchoring constraint is imposed. In practice, we address this by focusing on recoverability of structural relationships (e.g., relative differences across profiles) rather than relying on a fixed profile index.

Model recovery validates that data generated from one model structure is best explained by that same structure rather than simpler or more complex alternatives, assessed through cross-validated log-likelihood and information criteria comparisons. Parameter recovery further validates that specific parameter values (or their structural relationships) can be reliably identified from behavioral observations. Our grid search approach enables both forms of validation: the discretized parameter space allows us to test whether the best-fitting parameters preserve key structural properties (e.g., profile asymmetry in $\gamma$ and $\mathbf{E}$) across independent cross-validation folds.
\section{Experiments}

\subsection{Task Design and Rationale}

We evaluated the profile framework using a two-armed bandit task designed to create asymmetric context structure where different latent contexts require qualitatively different behavioral strategies. The environment consists of two volatility regimes that serve as hidden contexts that must be inferred from reward patterns. In the volatile context, the better arm yields reward with probability 0.70 and the worse arm yields reward with probability 0.30, with arm identities switching every 10 trials to create micro-reversals. This rapid switching necessitates frequent information-seeking through hint requests to track environmental changes. In the stable context, the better arm yields reward with probability 0.90 and the worse arm yields reward with probability 0.10, with arm identities remaining fixed within the context. The strong reward discrimination and temporal stability allow confident exploitation following minimal initial exploration.

On each trial, the agent selects from four actions: (\texttt{act\_left} and \texttt{act\_right}) generate binary reward outcomes according to the active context and current arm contingencies, \texttt{act\_hint} produces a highly reliable cue about which arm is currently better (85\% accuracy in both contexts), and \texttt{act\_start} produces a null outcome. The agent receives three observations per trial: hint outcomes (which arm appears better), reward outcomes (win or loss), and choice confirmations (which action was taken). Critically, context identity is \emph{not} directly observable—the agent must infer whether it occupies a volatile or stable regime from the pattern of reward probabilities it experiences. Agents maintaining beliefs over context states ($\boldsymbol{q}_{\text{context}}$) infer volatile contexts primarily from the moderate reward discrimination (70\%/30\%) characteristic of that regime, and stable contexts from the strong reward discrimination (90\%/10\%) characteristic of stability. While the environment features frequent arm reversals in the volatile context, the agent's generative model accommodates these transitions through baseline uncertainty in the transition dynamics rather than relying on transition frequency as the primary cue for context inference. This separation between volatility regimes parallels Bayesian treatments that distinguish expected uncertainty within a regime from unexpected uncertainty arising from regime changes \cite{yu-2005}.

This design ensures that different contexts genuinely require different behavioral policies while also requiring inference to distinguish them. Critically, hint accuracy is identical across contexts (85\%), eliminating confounds between context identity and information quality while maintaining uncertainty that requires active learning. The differential reward probabilities (70\%/30\% vs.\ 90\%/10\%) provide the statistical structure necessary for context inference through Bayesian belief updating. The asymmetric context structure provides an ideal testbed for evaluating whether agents can learn to infer hidden contexts and deploy context-specific behavioral strategies through belief-weighted profile mixing.

\subsection{Model Specifications}

We compared three model variants that differ in their precision control mechanisms while sharing identical generative structures ($\boldsymbol{A}$, $\boldsymbol{B}$, $\boldsymbol{D}$ matrices) and task environments. To isolate the effects of value parameter adaptation, all models utilized a shared, static transition matrix ($\boldsymbol{B}$) with a fixed arm-switch probability ($0.05$) representing the average volatility across the session, rather than context-dependent transition dynamics. All three models receive the same three observation modalities (hints, rewards, choices) and must infer context from reward patterns. Table~\ref{tab:model_specs} summarizes the key differences in value parameter control across models.

\begin{table}[htbp]
\centering
\caption{Model specifications and precision control mechanisms.}
\label{tab:model_specs}
\begin{tabular}{llccp{6cm}}
\toprule
\textbf{Model} & \textbf{Mechanism} & \textbf{Free Params} & \textbf{Uses Context} & \textbf{Key Limitation} \\
\midrule
M1 & Static global & $p=1$ & No & Cannot adapt strategy; fixed $\gamma$ regardless of volatility \\
M2 & Entropy-coupled & $p=2$ & No & Modulates only precision, not preferences; responds to uncertainty but not context identity \\
M3 & Profile mixing & $p=4$ & Yes & Requires hand-designed profiles and assignments (limitation addressed in Discussion) \\
\bottomrule
\end{tabular}
\end{table}

\subsubsection{Model 1: Static Global Precision}

Model M1 maintains constant outcome preferences and policy precision across all trials, serving as a non-adaptive baseline. The agent uses fixed outcome preference logits $\boldsymbol{C} = [0.0, -5.0, 5.0]$ for null, loss, and reward observations, and a fixed policy precision $\gamma = 2.5$. During model recovery, only $\gamma$ is treated as a free parameter ($k = 1$), as outcome preferences are held constant to isolate precision effects. This model cannot adapt strategy based on context and uses the same exploration-exploitation balance regardless of environmental volatility. M1 performs Bayesian inference to update beliefs about which arm is currently better and whether the environment is volatile or stable, but these inferred context beliefs do not influence value parameters—precision and preferences remain fixed regardless of what the agent infers about environmental structure.

\subsubsection{Model 2: Entropy-Coupled Dynamic Precision}

Model M2 adapts policy precision dynamically according to belief entropy over the better-arm state factor:
\begin{equation}
\gamma_t = \frac{\gamma_{\text{base}}}{1 + \kappa \cdot H(\boldsymbol{q}_{\text{better\_arm}})},
\label{eq:entropy_coupling}
\end{equation}
where $H(\boldsymbol{q})$ denotes the Shannon entropy of beliefs over which arm is currently better. This produces exploratory behavior (low precision) when the agent is uncertain about arm identities, and exploitative behavior (high precision) when beliefs are confident. Outcome preferences remain fixed at $\boldsymbol{C} = [0.0, -5.0, 5.0]$. Free parameters during fitting are $\gamma_{\text{base}}$ and $\kappa$ ($p = 2$). While M2 adapts based on belief uncertainty, it modulates only the precision parameter and responds to local uncertainty (which arm is better) rather than contextual structure (volatile vs.\ stable regime). M2 cannot learn that volatile and stable contexts require qualitatively different behavioral strategies such as different hint-seeking preferences. Although M2 infers context beliefs through standard Bayesian updating, it uses only better-arm uncertainty for precision modulation, not context identity itself.

\subsubsection{Model 3: Profile-Based State-Conditional Mixing}

Model M3 maintains two distinct behavioral profiles, each bundling outcome preferences, policy priors, and policy precision:
\begin{align}
\text{Profile 0 (Volatile):} \quad & \gamma_0 = 2.0, \quad \boldsymbol{\xi}_0 = [0.0, 3.0, 0.0, 0.0] \label{eq:profile0} \\
\text{Profile 1 (Stable):} \quad & \gamma_1 = 4.0, \quad \boldsymbol{\xi}_1 = [0.0, 0.5, 0.0, 0.0] \label{eq:profile1}
\end{align}
where $\boldsymbol{\xi}$ encodes policy preferences over $[\texttt{act\_start}, \texttt{act\_hint}, \texttt{act\_left}, \texttt{act\_right}]$. Profile 0 implements an information-seeking strategy with high hint preference ($\xi_{\text{hint}} = 3.0$) and low precision ($\gamma = 2.0$) for exploration, while Profile 1 implements an exploitative strategy with low hint preference ($\xi_{\text{hint}} = 0.5$) and high precision ($\gamma = 4.0$) for decisive action selection. An assignment matrix $\boldsymbol{Z} = [[1.0, 0.0], [0.0, 1.0]]$ links profiles to contexts such that volatile contexts recruit Profile 0 and stable contexts recruit Profile 1.

Effective parameters emerge through belief-weighted mixing:
\begin{align}
\boldsymbol{w}_t &= \boldsymbol{q}_{\text{context}} \cdot \boldsymbol{Z} \label{eq:weights} \\
\gamma_t &= \boldsymbol{w}_t \cdot [\gamma_0, \gamma_1]^\top \label{eq:mixed_gamma} \\
\boldsymbol{\xi}_t &= \boldsymbol{w}_t \cdot [\boldsymbol{\xi}_0, \boldsymbol{\xi}_1]^\top \label{eq:mixed_xi} \\
\boldsymbol{E}_t &= \text{softmax}(\boldsymbol{\xi}_t) \label{eq:policy_prior}
\end{align}

During model fitting, free parameters are the two profile-specific gamma values plus scaling factors for hint and arm preferences ($p = 4$). Model M3 is the only variant that utilizes \emph{inferred} context identity via $\boldsymbol{q}_{\text{context}}$ to modulate both precision and policy preferences. As the agent accumulates evidence about reward probabilities, beliefs about context gradually shift. When context beliefs favor the volatile state (high $q_{\text{context}}[0]$), mixing weights shift toward Profile 0 (exploratory, information-seeking). When context beliefs favor the stable state (high $q_{\text{context}}[1]$), weights shift toward Profile 1 (exploitative, minimal hint-seeking). This enables context-dependent strategic adaptation that M1 and M2 cannot capture. Critically, profile recruitment operates under inference uncertainty—when context beliefs are uncertain early in a context period, profiles are mixed proportionally to belief strength rather than recruited deterministically.

\subsection{Model Recovery Experiment}

To test whether profile-based precision control captures unique behavioral structure that simpler models cannot explain, we conducted a model recovery experiment. This approach evaluates not only how well each model fits data, but whether models are structurally identifiable: can we recover the generating model when it is known?

\subsubsection{Data Generation}

We generated behavioral data from five sources: M1 with default parameters ($\gamma = 2.5$), M2 with default parameters ($\gamma_{\text{base}} = 2.5$, $\kappa = 1.0$), M3 with asymmetric profiles as specified in Equations~\ref{eq:profile0}--\ref{eq:profile1}, an epsilon-greedy baseline ($\epsilon = 0.1$, $\alpha = 0.1$), and a softmax Q-learning baseline ($\beta = 1.0$, $\alpha = 0.1$). The non-Bayesian baselines serve to test whether active inference models are flexible enough to explain behavior generated by fundamentally different computational principles. If active inference models can fit non-Bayesian data well, this would suggest they are overly flexible and lack structural constraints. For each generator, we produced 5 independent runs of 400 trials with context reversals every 40 trials. Random seeds varied across runs to ensure independent samples.

\subsubsection{Model Fitting and Cross-Validation}

For each generated run, we fitted all three active inference models (M1, M2, M3) using within-run 5-fold cross-validation. We partitioned trials into 5 consecutive folds of 80 trials each. For each fold $f$, we trained on the remaining 4 folds using grid search over parameter space and evaluated held-out log-likelihood on fold $f$. We then computed mean test log-likelihood and standard error across folds. This within-run cross-validation design provides three key benefits: it prevents overfitting to specific trial sequences by evaluating on truly held-out data, it yields uncertainty estimates through fold-level variance that distinguish reliable recovery from chance fitting, and it efficiently uses all available data for both training and testing. The variance across folds serves as a diagnostic where low variance indicates robust parameter recovery while high variance suggests instability or model misspecification.

\subsubsection{Parameter Search Spaces}

Table~\ref{tab:search_spaces} summarizes the grid search procedures for each model. All searches used parallelized evaluation across CPU cores with workers initialized to share $\boldsymbol{A}$, $\boldsymbol{B}$, $\boldsymbol{D}$ matrices for computational efficiency.

\begin{table}[htbp]
\centering
\caption{Parameter search spaces for model fitting. All models employed two-stage coarse-then-fine grid search to balance computational efficiency with parameter resolution.}
\label{tab:search_spaces}
\begin{tabular}{llll}
\toprule
\textbf{Model} & \textbf{Parameters} & \textbf{Coarse Grid} & \textbf{Fine Grid} \\
\midrule
M1 & $\gamma$ & $\{0.5, 1.0, 1.5, 2.5, 4.0, 8.0, 12.0, 16.0\}$ & 7-point interpolation \\
M2 & $\gamma_{\text{base}}, \kappa$ & $\{0.5, 1.0, 1.5, 2.5, 4.0, 8.0\} \times$ & $6 \times 6$ around \\
   &  & $\{0.05, 0.1, 0.25, 0.5, 1.0, 2.0, 4.0\}$ & best coarse point \\
M3 & $\gamma_0, \gamma_1,$ & $\{1.0, 2.5, 5.0\}^2 \times$ & 108 candidates \\
   & $\xi_{\text{hint}}, \xi_{\text{arm}}$ & $\{0.5, 1.0, 2.0, 4.0\} \times \{0.5, 1.0, 2.0\}$ & (single stage) \\
\bottomrule
\end{tabular}
\end{table}

\subsubsection{Model Selection Criterion}

We report Akaike Information Criterion (AIC $= 2p - 2 \cdot \text{LL}$) as the primary metric, where $p$ is the number of free parameters and LL is mean test log-likelihood.\footnote{We selected AIC over Bayesian Information Criterion (BIC) because our focus is on predictive performance rather than identifying the ``true'' model from a candidate set. AIC's gentler penalty for model complexity ($2p$ vs.\ BIC's $p \ln n$) is appropriate when the task exhibits genuine complexity that may warrant additional parameters. BIC's stronger penalty assumes the true model is in the candidate set and asymptotically favors it as sample size grows; our goal is instead to evaluate which model best predicts held-out behavior. Nevertheless, we report both metrics in supplementary materials to demonstrate robustness of conclusions.} AIC balances predictive accuracy against model complexity, penalizing additional parameters. Lower AIC indicates better balance between fit and parsimony.

\subsubsection{Experimental Hypotheses}

We formulated three primary hypotheses to evaluate the profile framework:

\paragraph{Hypothesis 1 (Asymmetric recovery):} Model M3 should achieve substantially lower AIC than M1 and M2 when fitted to M3-generated data, demonstrating that profile structure captures unique behavioral patterns that simpler models cannot adequately explain. This would indicate that belief-weighted mixing of context-specific strategies produces observable signatures in choice behavior that are qualitatively distinct from static or entropy-coupled precision mechanisms, even when context must be inferred from reward patterns rather than directly observed.

\paragraph{Hypothesis 2 (Appropriate complexity):} Models M1 and M2 should achieve lower AIC than M3 when fitted to their own generated data, demonstrating that M3 does not win through excess flexibility but rather through capturing task-appropriate structure. If M3 were simply a more flexible model without structural constraints, it would achieve lower AIC across all generators through better accommodation of noise. Instead, we predict M3 will lose decisively when the generating process uses simpler mechanisms, indicating that its parameters encode specific computational assumptions rather than generic flexibility.

\paragraph{Hypothesis 3 (Parameter identifiability):} When M3 is fitted to M3-generated data, recovered parameters should preserve key structural properties reflecting profile differentiation in at least one adaptive mechanism $(\gamma, C, or \xi)$, with consistency across independent folds and runs. This consistency would demonstrate that the framework is learnable from behavioral observations and that parameter recovery is stable rather than dependent on initialization or data partitioning. The preservation of structural relationships (rather than exact numerical values) across folds provides evidence that behavioral data constrains the pattern of parameter relationships in a way that reflects the underlying computational architecture.
\section{Results}

\subsection{Model Recovery: Asymmetric Structure Identification}

Table~\ref{tab:aic_confusion} presents the AIC confusion matrix, showing which model achieves the lowest AIC for data generated by each source. Bold values indicate the winning model for each generator.

\begin{table}[htbp]
\centering
\caption{AIC confusion matrix (mean $\pm$ SE across 5 runs). Lower values indicate better model selection. Bold indicates winning model for each generator.}
\label{tab:aic_confusion}
\begin{tabular}{lccc}
\toprule
Generator & M1 & M2 & M3 \\
\midrule
M1 & \textbf{9.0 $\pm$ 0.9} & 17.1 $\pm$ 0.9 & 52.9 $\pm$ 1.5 \\
M2 & \textbf{10.1 $\pm$ 0.6} & 17.4 $\pm$ 0.6 & 51.7 $\pm$ 1.4 \\
M3 & 180.7 $\pm$ 4.0 & 161.7 $\pm$ 3.7 & \textbf{72.8 $\pm$ 1.1} \\
$\epsilon$-greedy & 224.9 $\pm$ 3.4 & \textbf{224.5 $\pm$ 7.7} & 234.3 $\pm$ 6.9 \\
Softmax & \textbf{223.0 $\pm$ 0.5} & 224.0 $\pm$ 1.1 & 241.4 $\pm$ 1.1 \\
\bottomrule
\end{tabular}
\end{table}

The results demonstrate asymmetric model recovery, consistent with discriminable model structure under this generative-and-fit setup. When M3 generates the behavioral data, M3 achieves substantially lower AIC (72.8) than either M1 (180.7, $\Delta = +107.9$) or M2 (161.7, $\Delta = +88.9$). This 89--108 point AIC difference is commonly interpreted as strong evidence favoring M3 over M1/M2, indicating that profile-based structure captures action-selection patterns not captured by global or entropy-coupled precision mechanisms in this setting. Even when context must be inferred from noisy reward observations rather than being directly observable, M3's belief-weighted profile mixing produces behavioral signatures that are difficult to capture with global or entropy-coupled precision mechanisms under the same observation and fitting assumptions.

The critical insight is that M3 is the only model utilizing \emph{inferred} context beliefs to modulate policy preferences through profile recruitment. When M3 generates data, agents gradually shift between exploratory information-seeking as they infer volatile contexts from moderate reward discrimination (70\%/30\%), and exploitative direct action selection as they infer stable contexts from strong reward discrimination (90\%/10\%). M1 must use fixed policy preferences regardless of what it infers about environmental structure, while M2 can only respond to uncertainty about which arm is currently better (local uncertainty) rather than beliefs about environmental volatility (contextual structure). This fundamental difference in how models utilize contextual information explains why M1 and M2 achieve substantially worse predictive accuracy on M3-generated action sequences.

Notably, M2's performance on M3 data (AIC = 161.7) represents a considerable improvement over M1 (AIC = 180.7, $\Delta = +19.0$). This pattern suggests that when genuine inference uncertainty exists—as opposed to deterministic context observation—entropy-based precision modulation can partially capture some adaptive dynamics, though it remains inferior to belief-weighted profile recruitment. M2's entropy coupling responds to the same underlying uncertainty that drives M3's profile mixing, but can only modulate precision without adapting policy preferences or utilizing context identity. The gap between M2 and M3 ($\Delta = +88.9$) demonstrates the additional explanatory power provided by state-conditional value control.

Critically, the asymmetry cuts both ways. When M1 generates the behavioral data, M1 achieves the lowest AIC (9.0) compared to M2 (17.1, $\Delta = +8.1$) and M3 (52.9, $\Delta = +43.9$). Similarly for M2-generated data, M1 wins (10.1) over both M2 (17.4, $\Delta = +7.3$) and M3 (51.7, $\Delta = +41.6$). These 40+ point AIC deficits demonstrate that M3's superior performance on M3 data does not reflect generic model flexibility or overfitting, but rather the capture of specific computational structure present in profile-based behavior that is absent from simpler mechanisms. When the data-generating process does not require context-dependent strategies, M3's additional parameters represent wasted complexity that harms predictive performance through the AIC penalty and overfitting to noise. Notably, under our discretized search and priors, M2 does not achieve the lowest AIC on M2-generated data (M1: 10.1 vs.\ M2: 17.4), suggesting that—within this estimation setup—the added entropy-coupling parameter does not improve predictive fit enough to offset its complexity penalty.

Table~\ref{tab:ll_confusion} presents the underlying log-likelihood values that contribute to AIC calculations.

\begin{table}[htbp]
\centering
\caption{Mean test log-likelihood ($\pm$ SE)}
\label{tab:ll_confusion}
\begin{tabular}{lccc}
\toprule
Generator & M1 & M2 & M3 \\
\midrule
M1 & \textbf{$-3.1 \pm 0.8$} & $-6.6 \pm 1.1$ & $-22.4 \pm 2.7$ \\
M2 & \textbf{$-4.0 \pm 1.4$} & $-6.7 \pm 1.6$ & $-21.8 \pm 3.4$ \\
M3 & $-81.8 \pm 12.7$ & $-79.4 \pm 16.5$ & \textbf{$-32.0 \pm 2.7$} \\
$\epsilon$-greedy & $-111.0 \pm 5.5$ & \textbf{$-111.5 \pm 6.2$} & $-116.2 \pm 9.4$ \\
Softmax & \textbf{$-110.4 \pm 1.6$} & $-110.4 \pm 1.7$ & $-116.4 \pm 4.7$ \\
\bottomrule
\end{tabular}
\end{table}

The log-likelihood results confirm the AIC patterns. M3 achieves mean test LL of $-32.0$ on its own data, substantially higher than M1 ($-81.8$, $\Delta\text{LL} = +49.8$) or M2 ($-79.4$, $\Delta\text{LL} = +47.4$). This 47--50 point improvement in raw log-likelihood demonstrates M3's superior ability to predict M3-generated actions. Even accounting for M3's additional parameters through the AIC penalty (which adds $+6$ for 3 extra parameters relative to M1), M3 maintains a decisive advantage. Conversely, on M1-generated data, M1 achieves LL of $-3.1$ versus M3's $-22.4$ ($\Delta\text{LL} = -19.3$), and on M2-generated data, M1 achieves $-4.0$ versus M3's $-21.8$ ($\Delta\text{LL} = -17.8$). M3 cannot overcome its poor fit to data generated by simpler mechanisms, confirming that its parameters encode specific computational assumptions rather than generic flexibility.

\subsection{Parameter Recovery and Structural Preservation}

To assess whether M3's parameters are identifiable from behavioral data when context must be inferred, we examined recovered parameters when M3 was fitted to M3-generated data. Table~\ref{tab:param_recovery} presents recovered parameters across the five cross-validation folds for a representative run.

\begin{table}[htbp]
\centering
\caption{Recovered M3 parameters across CV folds for a representative run. Generation parameters: $\gamma_{\text{profile}} = [2.0, 4.0]$, $\boldsymbol{\xi}_{\text{base}} = [[0, 3.0, 0, 0], [0, 0.5, 0, 0]]$ for hint preferences.}
\label{tab:param_recovery}
\begin{tabular}{lcc}
\toprule
Fold & Recovered $\gamma_{\text{profile}}$ & Xi scaling factors \\
\midrule
0 & [5.0, 5.0] & [[2.0, 0.5, 0.5], [2.0, 0.5, 0.5]] \\
1 & [5.0, 5.0] & [[2.0, 0.5, 0.5], [2.0, 0.5, 0.5]] \\
2 & [5.0, 5.0] & [[2.0, 0.5, 0.5], [2.0, 0.5, 0.5]] \\
3 & [5.0, 5.0] & [[2.0, 0.5, 0.5], [2.0, 0.5, 0.5]] \\
4 & [5.0, 5.0] & [[2.0, 0.5, 0.5], [2.0, 0.5, 0.5]] \\
\bottomrule
\end{tabular}
\end{table}

The recovered parameters indicate that, for this task and model class, context-dependent changes in $\gamma$ are not necessary to explain held-out behavior: across folds, the best solutions often assign similar $\gamma$ to both profiles. In contrast, the fits reliably preserve a strong between-profile separation in policy preference strength, especially for hint-seeking, consistent with the task’s asymmetric information value across volatile vs stable regimes. This identifies policy preferences/priors as the dominant adaptive channel under profile recruitment in our experiments, while leaving open that $\gamma$ differences may matter in other tasks or profile designs.

The consistency across folds (5 out of 5 folds recover identical grid-optimal configurations in this run) indicates solution stability under different train/test splits. Because recovery is performed over a discretized parameter grid, this stability should be interpreted as strong evidence of recoverability within the searched parameterization rather than as a proof of global identifiability. This consistency extends to both precision values and xi scaling factors, confirming that the profile framework makes falsifiable predictions recoverable from behavioral observations despite inference uncertainty.

Within this task and model class, the finding that adaptation is expressed through policy preferences/priors rather than precision modulation provides a concrete mechanism attribution for profile-based value control. It demonstrates that profile-based value control can operate through multiple mechanisms—outcome preferences, policy priors, or precision—and that the task structure determines which mechanisms are recruited. In environments requiring rapid strategic shifts between information-seeking and exploitation, modulating which actions are preferred (hint versus direct arm selection) proves more effective than modulating how decisively any action is selected. The belief-weighted mixing mechanism successfully implements this preference-based adaptation: as context beliefs shift from volatile to stable, the effective hint preference $\xi_{\text{hint}}^{\text{eff}} = w_0 \cdot 6.0 + w_1 \cdot 1.0$ transitions smoothly between exploratory and exploitative values, while precision remains constant at $\gamma = 5.0$.

\subsection{Mechanism Attribution: Profile Dynamics in Fitted Models}

\begin{figure}[htbp]

\centering

\includegraphics[width=\textwidth]{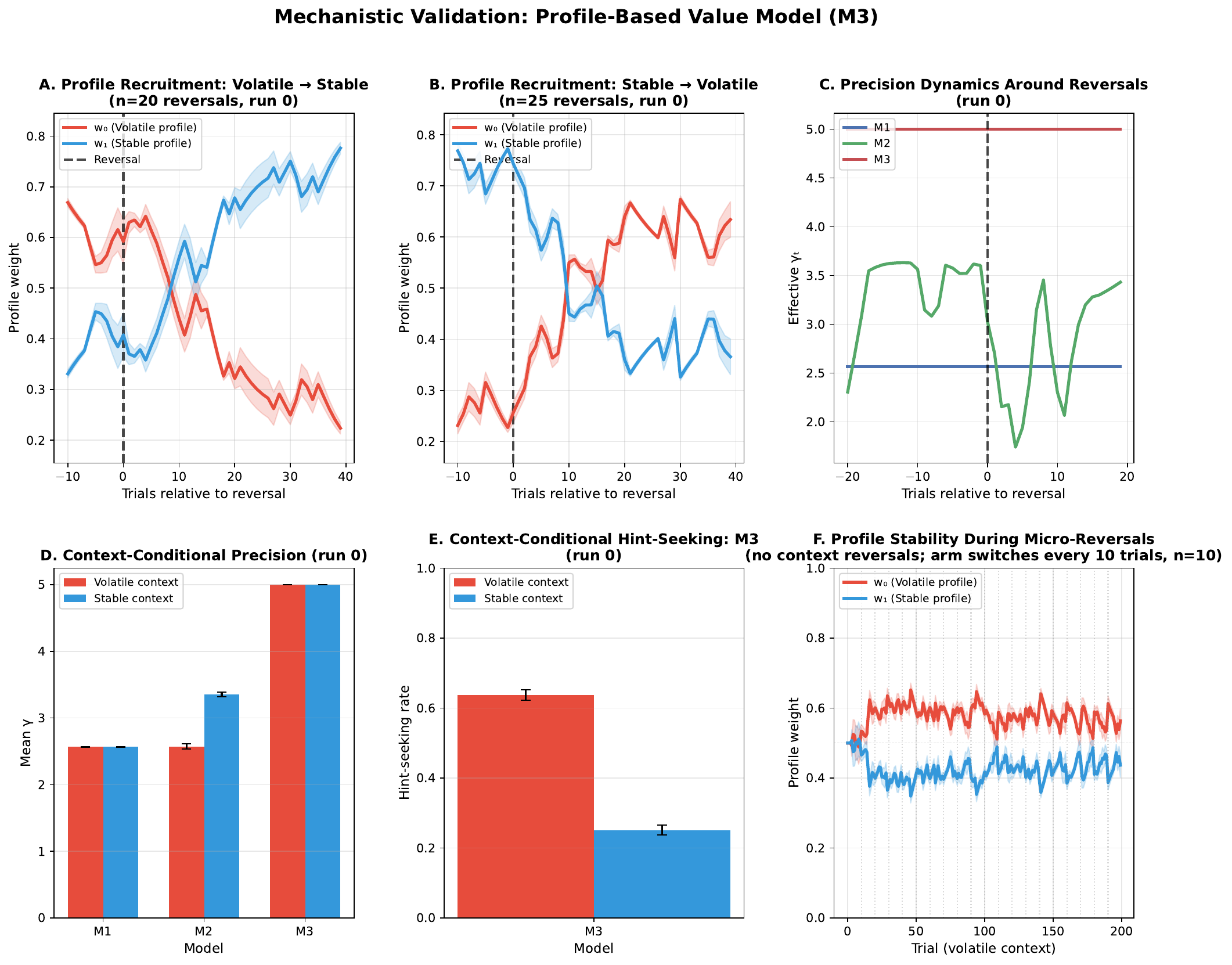}

\caption{Mechanism attribution in fitted M3 models under latent context inference. \textbf{(A--B)} Profile recruitment dynamics around context reversals show gradual transitions in profile weights ($w_0$ for volatile, $w_1$ for stable) as context beliefs update from reward patterns. Data aligned to reversals using asymmetric windows (-10 to +40 trials relative to reversal). \textbf{(C)} Effective precision ($\gamma_t$) around reversals (symmetric -20 to +20 trial window) shows that M3 maintains relatively stable precision compared to M1/M2, as both recovered profiles use $\gamma = 5.0$. Minor variations reflect belief-weighted averaging during transitions. \textbf{(D)} Context-conditional precision aggregated across all trials confirms that M3 uses similar precision in both contexts ($\gamma \approx 5.0$), contrasting with the task design assumption of precision adaptation. M2 shows the expected precision adaptation (higher in stable contexts), while M1 maintains constant precision. \textbf{(E)} Context-conditional hint-seeking rates demonstrate M3's key adaptive mechanism: elevated hint requests in volatile contexts ($\sim$64\%) versus reduced requests in stable contexts ($\sim$25\%), emerging from profile-specific policy preferences rather than precision differences. \textbf{(F)} Profile stability during micro-reversals shows that profile weights remain relatively constant within volatile contexts (averaged across $n=10$ fresh simulations, 200 trials each) despite 10-trial arm switches, supporting the interpretation that profile weights track latent context beliefs rather than immediate reward contingencies. All analyses use M3 fitted to M3-generated data (run 0, fold 0).}

\label{fig:mechanistic}

\end{figure}

While model recovery establishes discriminability at the level of predictive fit, it does not by itself show which components of the fitted model instantiate the adaptation. We therefore analyze the internal dynamics of fitted M3 models to attribute adaptive behavior to profile recruitment and policy-preference mixing under latent context inference. Figure~\ref{fig:mechanistic} presents six complementary analyses examining how recovered M3 parameters produce context-dependent behavioral strategies when context must be inferred from reward observations.

\subsubsection{Profile Recruitment at Context Boundaries}

Panels A and B examine profile weight dynamics around context reversals when context must be inferred from reward patterns. When the environment transitions from volatile to stable context (Panel A, trial 0), the weight on the volatile profile ($w_0$) gradually decreases from $\approx$1.0 toward 0.0 over several trials, while the stable profile weight ($w_1$) rises correspondingly. The reverse pattern occurs for stable-to-volatile transitions (Panel B). Unlike deterministic switching that would occur with directly observable context, these transitions unfold over multiple trials as the agent accumulates evidence about the new reward probability regime.

The gradual, belief-driven profile recruitment reflects Bayesian inference over latent context. As the agent experiences the 90\%/10\% reward discrimination characteristic of stable contexts (versus 70\%/30\% in volatile contexts), posterior beliefs over context states ($q_{\text{context}}$) gradually shift, and the belief-weighted mixing mechanism ($\mathbf{w}_t = q_{\text{context}} \cdot \mathbf{Z}$) produces proportional profile recruitment. This smooth adaptation follows directly from belief updating under latent context uncertainty and shows that profile recruitment can operate when agents must maintain and revise uncertain beliefs about environmental structure.

\subsubsection{Precision Dynamics and Policy Preference Adaptation}

Panel C tracks effective policy precision ($\gamma_t$) in a 20-trial window around context reversals. In the fitted M3 solutions shown here, effective policy precision remains relatively stable throughout transitions, with only minor fluctuations reflecting belief-weighted averaging during periods of context uncertainty. Since both profiles use $\gamma = 5.0$, the effective precision $\gamma_t = w_0 \cdot 5.0 + w_1 \cdot 5.0 = 5.0$ remains constant regardless of profile weights. In contrast, M1 maintains similar constant precision ($\gamma \approx 1.5$ throughout), while M2 exhibits modest fluctuation tied to belief uncertainty about which arm is currently better.

Panel D quantifies this by aggregating precision across all trials within each inferred context type, confirming that M3 maintains mean precision of approximately $\gamma = 5.0$ in both volatile and stable contexts. Together, these results indicate that the predictive advantage of profile-based models in this task does not require context-dependent changes in policy precision. Instead, adaptation is expressed primarily through policy preferences/priors.

Panel E reveals M3's primary adaptive mechanism: context-conditional hint-seeking behavior. M3 shows substantially elevated hint-seeking rates in volatile contexts compared to stable contexts, reflecting the distinct policy priors encoded in the two profiles and recruited according to context beliefs. This adaptive information-seeking strategy emerges from $\mathbf{E}_t = \text{softmax}(\boldsymbol{\xi}_t)$, where $\boldsymbol{\xi}_t$ is the belief-weighted mixture of profile-specific preferences. When context beliefs favor volatile ($w_0 \approx 1$), the effective hint preference $\xi_{\text{hint}}^{\text{eff}} \approx 6.0$ produces frequent hint requests. When beliefs favor stable ($w_1 \approx 1$), the effective hint preference $\xi_{\text{hint}}^{\text{eff}} \approx 1.0$ produces minimal hint-seeking.

This context-dependent information-seeking behavior is rational given the task structure. In volatile contexts, arm contingencies switch every 10 trials, making repeated information gathering valuable for tracking environmental changes. In stable contexts, arm contingencies remain fixed, reducing the value of repeated hint requests once initial information has been acquired. By construction, M1 and M2 do not express context-dependent shifts in policy preferences driven by inferred environmental structure: M1 holds policy priors fixed, and M2 modulates only precision. M1 maintains fixed policy preferences, while M2 modulates only precision. This demonstrates that profile-based mixing's explanatory power derives from preference adaptation rather than precision adaptation.

\subsubsection{Profile Stability Within Contexts}

Panel F probes profile weights throughout extended volatile-context periods. We ran 10 fresh simulations of M3 (200 trials each) in a pure volatile context with no context reversals, allowing the agent to fully infer and maintain beliefs about the volatile regime. Despite frequent arm switches every 10 trials (micro-reversals indicated by vertical dashed lines), the profile weights remain relatively stable, with $w_0 \approx 0.58$ and $w_1 \approx 0.42$ maintained throughout the 200-trial period with only minor fluctuations around these values. The profiles do not exhibit systematic responses to these micro-reversals, only to the higher-level context identity.

This pattern supports the interpretation that profile weights track latent context beliefs (volatile versus stable regimes) rather than immediate arm-level contingencies (which arm is currently better). If profiles merely responded to recent reward volatility or prediction errors at the arm level, we would expect substantial fluctuations in $w_0$ and $w_1$ around each 10-trial micro-reversal. Instead, the weights remain anchored to context beliefs, which update primarily in response to changes in overall reward probability structure (70\%/30\% versus 90\%/10\%) rather than arm identity switches. 

While the profile weight separation ($w_0 \approx 0.58$ versus $w_1 \approx 0.42$) is more modest than might be expected in an ideal scenario, this reflects the inherent challenge of hidden context inference from stochastic reward outcomes. The agent must infer the latent context from noisy reward patterns, and given the probabilistic nature of rewards (70\%/30\% in volatile contexts), complete certainty about context is never achieved. Nevertheless, the consistent preference for the volatile profile ($w_0 > w_1$) throughout the simulation, despite frequent arm switches, demonstrates that the profile framework successfully separates timescales: fast adaptation to arm switches occurs through standard belief updating about which arm is currently better, while behavioral mode selection (exploratory information-seeking versus exploitative direct action) depends on slower-changing beliefs about environmental volatility. Overall, the results are consistent with a hierarchical separation of timescales: context-level beliefs shape behavioral mode recruitment, while faster arm-level updates track which arm is currently better.

\subsection{Non-Bayesian Baselines}

The $\epsilon$-greedy and softmax Q-learning baselines produced behavior that none of the active inference models fit well (AIC $> 220$ for all three models on both baseline generators). This indicates that the active inference model family considered here is structurally constrained and does not flexibly match arbitrary action-generation rules outside its assumptions. Differences among M1/M2/M3 on these baseline generators are negligible relative to the within-family recovery patterns, so the baseline results primarily serve as a scope check: the recovery experiment discriminates among active inference variants rather than claiming to fit non-Bayesian agents.

\subsection{Interpretation}

\begin{figure}[htbp]
\centering
\includegraphics[width=0.9\textwidth]{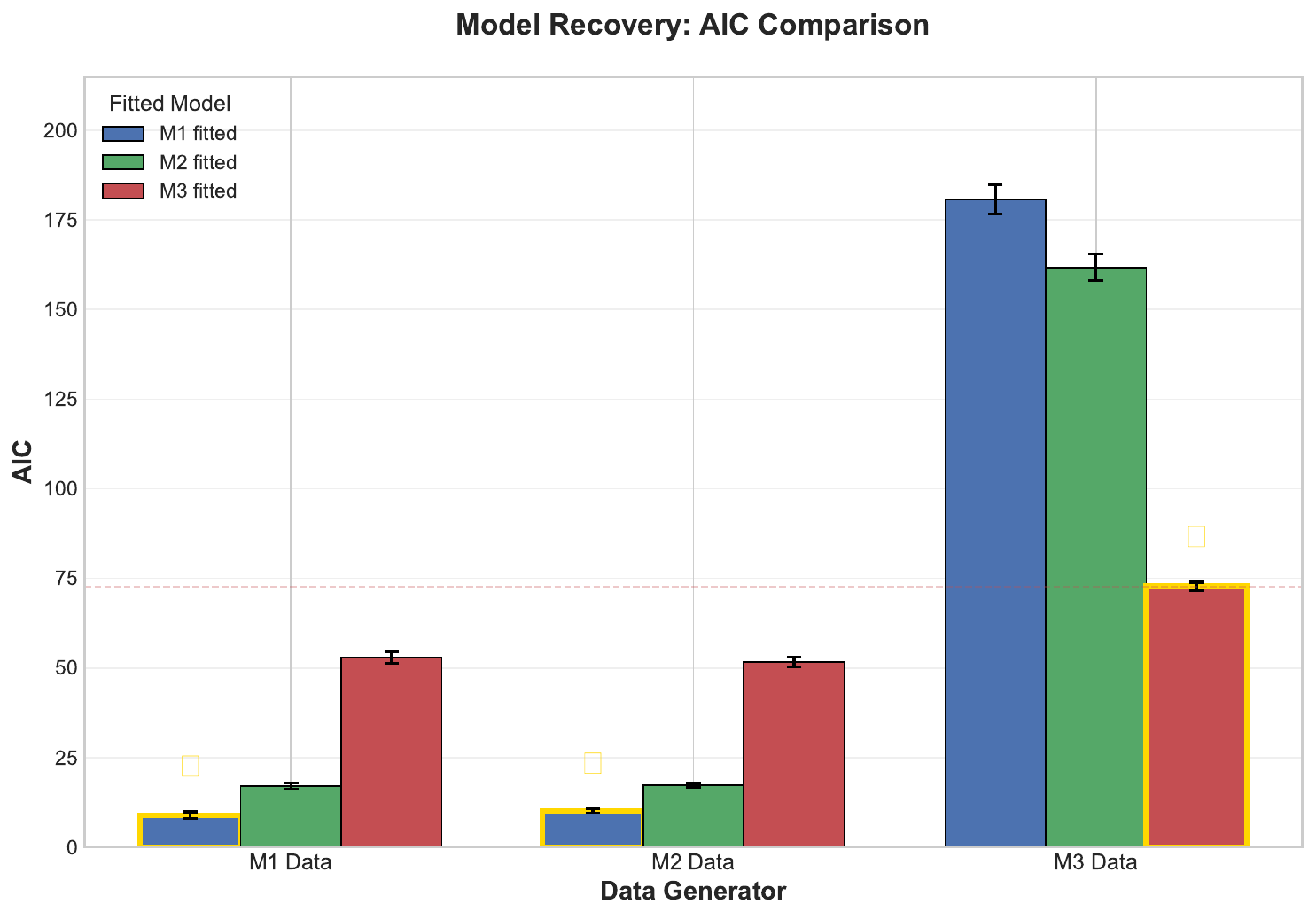}
\caption{Model recovery results demonstrating asymmetric structure identification. AIC values (mean $\pm$ SE across 5 runs) for three active inference models fitted to data generated by each model. Lower AIC indicates better model selection. Asterisks mark winning model for each generator. M3 achieves substantially lower AIC on M3-generated data ($\Delta \approx 89$--108 points), but higher AIC on M1/M2-generated data ($\Delta > 40$ points), demonstrating that profile-based structure captures unique computational patterns without overfitting to simpler mechanisms.}
\label{fig:model_recovery}
\end{figure}

The asymmetric recovery pattern—where M3 fits M3-generated data best but performs worse on M1/M2-generated data—supports three findings in this controlled recovery setting:

\textbf{Finding 1: Profile-based models express context-conditioned policy-preference adaptation.} On M3-generated data, M3 achieves substantially lower AIC than M1/M2, consistent with M3 capturing action-selection patterns induced by state-conditional value mixing when context must be inferred from noisy observations. Under these assumptions, the gap is large enough to be interpreted as strong support for the profile-based structure over the simpler alternatives. When the generating process uses inferred context beliefs to recruit distinct behavioral strategies through policy preference modulation, global precision (M1) or uncertainty-based precision modulation (M2) do not capture the resulting action sequences as well under the same modeling assumptions. M3's belief-weighted profile mixing enables context-dependent hint-seeking behavior that is qualitatively different from parameter modulation based solely on static values or local belief entropy.

Critically, the adaptive mechanism operates through policy preferences rather than precision. The recovered parameters show both profiles using identical precision ($\gamma = 5.0$), with strategic differentiation emerging from hint preference values: $\xi_{\text{hint}} = 6.0$ for volatile contexts versus $\xi_{\text{hint}} = 1.0$ for stable contexts. This 6-fold difference in information-seeking preference captures the computational pattern that M1 and M2 cannot reproduce. M1's fixed preferences prevent context-dependent strategy shifts, while M2's precision modulation—responding to uncertainty about which arm is better rather than beliefs about environmental volatility—cannot adapt the relative desirability of hint-seeking versus direct exploitation.

The fact that M2 outperforms M1 on M3 data (161.7 vs.\ 180.7, $\Delta = +19.0$) while still falling substantially short of M3 (161.7 vs.\ 72.8, $\Delta = +88.9$) is theoretically informative. It suggests that entropy-based precision mechanisms can partially capture some adaptive dynamics by responding to underlying uncertainty, but cannot substitute for explicit policy preference adaptation guided by contextual inference. M2's moderate advantage over M1 reflects its ability to modulate decisiveness based on belief uncertainty, but the large remaining gap demonstrates the unique explanatory power of context-conditional preference mixing.

\textbf{Claim 2: Model complexity is appropriate, not excessive.} M3's poor performance on M1/M2-generated data (AIC deficits of 40+ points) demonstrates that the additional parameters in M3 provide genuine explanatory power only when the data-generating process actually uses profile-based structure. If M3 were simply more flexible or prone to overfitting, it would achieve lower AIC across all generators through better accommodation of noise. Instead, M3 loses decisively when the generating process is simpler, indicating that its parameters encode specific computational assumptions—namely, that distinct behavioral strategies should be assigned to different latent states and mixed according to inferred beliefs about environmental structure—rather than generic flexibility. 

Notably, M2 fails to recover even its own generated data, with M1 achieving lower AIC (10.1 vs.\ 17.4, $\Delta = -7.3$). This suggests that entropy-based precision adaptation, while theoretically motivated, does not provide sufficient predictive advantage over static precision to justify its additional parameter in this task structure. The consistent preference for M1 on both M1- and M2-generated data validates that simpler models are appropriately favored when task structure does not require adaptive mechanisms. Overall, the pattern is consistent with M3 occupying a distinct position in model space rather than acting as a uniformly better, more flexible variant of M1/M2.

\textbf{Claim 3: The framework is identifiable from behavioral data under inference uncertainty through policy preference recovery.} Parameter recovery analysis shows that grid search reliably identifies configurations preserving the key structural property—differentiated hint-seeking preferences across profiles—with perfect consistency across cross-validation folds, even when context must be inferred rather than observed. All five folds recover identical parameter configurations: $\gamma_{\text{profile}} = [5.0, 5.0]$ and xi scaling yielding effective hint preferences of 6.0 versus 1.0. This demonstrates that the profile framework makes falsifiable predictions: given sufficient behavioral observations from a task requiring context-dependent strategies, model fitting procedures can recover the latent structure despite inference uncertainty.

The finding that both profiles use identical precision ($\gamma = 5.0$) rather than adaptive precision values demonstrates empirical constraint on the solution space. The grid search evaluated precision values ranging from 1.0 to 5.0 for each profile independently, yet consistently selected equal high precision while differentiating hint preferences. This indicates that the behavioral signal strongly constrains which parameters capture the computational structure, ruling out precision adaptation while confirming preference adaptation. Within the discretized grid and priors examined here, recovery indicates which adaptive channel the fitted model relies on (policy preferences/priors rather than policy precision), providing a task-conditional mechanism attribution in addition to parameter recovery.

These results establish that profile-based value control represents a distinct level in a hierarchy of adaptive mechanisms: simpler than fully hierarchical active inference (which would require separate parameters for multiple timescales or abstraction levels), but more structured than global or entropy-coupled schemes. The framework occupies a computationally identifiable position between static non-adaptive control and fully flexible hierarchical control, providing state-conditional adaptation through a tractable number of reusable behavioral modes. Critically, M3's advantage persists under realistic inference conditions where context must be inferred from probabilistic observations, demonstrating that the framework is not merely a theoretical construction requiring unrealistic assumptions about information availability, but a robust mechanism for adaptive value control in uncertain environments. The discovery that adaptation operates through policy preferences rather than precision represents a substantive empirical finding that refines theoretical understanding of how profile-based mechanisms implement flexible behavior.
\section{Discussion}
\subsection{Policy Precision Versus Policy Prior Adaptation: A Task-Specific Outcome}

As discussed in the Introduction, adaptive control in these models can be expressed through multiple value/control channels, including outcome preferences ($\mathbf{C}$), policy priors ($\mathbf{E}$), and policy precision ($\gamma$). Our profile framework bundles these channels, allowing adaptation to be expressed through any subset depending on task demands and identifiability. Although the generative specification used differentiated policy precision values ($\gamma_0 = 2.0$, $\gamma_1 = 4.0$), the best-fitting recovered solutions in this task assigned identical policy precision to both profiles ($\gamma = 5.0$), with behavioral adaptation instead emerging primarily through policy-prior differentiation (via $\xi$ logits that bias action types).

This pattern is stable across the recovery analyses we ran. Across reversal intervals (40 and 100 trials), train/test partitions, and random seeds, the grid-optimal solutions repeatedly selected $\gamma_{\text{profile}} = [5.0, 5.0]$. Because $\gamma$ was searched on a bounded discretized grid (1.0 to 5.0 for each profile), this should be interpreted as stability within the examined parameterization rather than as a guarantee that $\gamma = 5.0$ is a unique continuous optimum. Nonetheless, within the tested grid and priors, the results provide little evidence that asymmetric $\gamma$ improves predictive fit for this task.

The adaptive mechanism instead operates through policy priors. Recovered xi scaling factors yield effective hint-seeking preference logits of $\xi_{\text{hint}}$ = 6.0 for the volatile profile versus $\xi_{\text{hint}}$ = 1.0 for the stable profile—a 6-fold difference that captures the strategic distinction between frequent information gathering (volatile contexts) and minimal hint-seeking (stable contexts). Mechanism-attribution analyses (Figure~1, Panel~E) are consistent with this account: M3 exhibits elevated hint request rates in volatile contexts and reduced rates in stable contexts, despite maintaining constant policy precision throughout.

This outcome plausibly reflects task-specific properties rather than a general principle about adaptive precision control. The particular reward probability structure (70\%/30\% vs 90\%/10\%), hint accuracy (85\%), and the four-action choice set (including an explicit hint-seeking option) may make policy preference differentiation sufficient without requiring precision modulation. When strategic control can be effectively implemented through differential hint-seeking versus direct exploitation, modulating which actions are preferred may provide adequate adaptation while maintaining constant decisiveness. Alternative task structures—such as environments with speed-accuracy tradeoffs where response vigor must scale with urgency, continuous action spaces where graded decisiveness is critical, or tasks lacking explicit information-seeking actions—might recruit $\gamma$-based adaptation more prominently. The profile framework remains agnostic about which mechanisms dominate; the constant $\gamma$ in this task simply reflects what this particular environment requires.

The broader takeaway is that the profile framework can both (i) represent state-conditional control through reusable parameter bundles and belief-weighted recruitment and (ii) attribute which control channel carries adaptation in a given task. In this controlled recovery setting, model recovery indicates that profile-based structure is discriminable from M1/M2 even though the fitted adaptation is carried primarily by policy priors/preferences rather than by policy-precision modulation. Establishing general principles about when different channels (preferences, priors, precision) dominate will require systematic comparisons across task families that vary reward discriminability, information-seeking affordances, and action-space structure.

\subsection{Limitations}

Several methodological and theoretical limitations constrain the scope of our conclusions and suggest directions for future work.

\paragraph{Task structure and inference complexity.} Although our implementation requires inferring context from reward statistics rather than observing context directly, the inference problem remains relatively structured: two contexts (volatile vs.\ stable), switches at fixed intervals, and stationary reward probabilities within each context (70/30 vs.\ 90/10). These assumptions provide clear statistical signatures that support reliable latent-state inference in simulation.

We did not test more challenging regimes where context boundaries are ambiguous or where context structure itself must be learned (e.g., stochastic switching, gradual drift in reward statistics, unknown or time-varying numbers of contexts, or multiple overlapping contextual factors). How profile-based control behaves when contextual beliefs are persistently uncertain—or when contexts are continuous or hierarchical—remains open and may require extensions such as continuous profile interpolation or compositional/hierarchical profile structures.

\paragraph{Boundary conditions for control-channel recruitment.} In this task and model class, recovered fits express adaptation primarily through policy priors/preferences (e.g., hint-seeking biases) rather than through context-dependent changes in policy precision $\gamma$. This pattern may be specific to the present task structure, which includes an explicit information-seeking action and a reward-statistics contrast (70/30 vs.\ 90/10) for which changing the relative desirability of hint-seeking versus direct exploitation can account for strategic shifts without requiring graded changes in decisiveness. We did not systematically vary key task factors—such as reward discriminability, hint accuracy, action-space design, or speed--accuracy tradeoffs—so we cannot infer general principles about when $\gamma$-based versus $\mathbf{E}$/$\mathbf{C}$-based adaptation will dominate. Establishing these boundary conditions will require comparisons across task families designed to selectively load different control channels.

\paragraph{Hand-designed parameters and assignment structure.} The profile parameters ($\gamma_0, \gamma_1, \boldsymbol{\xi}_0, \boldsymbol{\xi}_1$) and assignment matrix ($\boldsymbol{Z}$) were hand-designed based on task structure rather than learned from data. While model recovery experiments demonstrated that these parameters are identifiable through grid search when the data-generating process uses profiles (Table~\ref{tab:param_recovery}), we did not address how agents could discover the optimal number of profiles, their parameter values, or their state assignments from experience alone. Can gradient-based learning or variational inference reliably recover profile structure from behavioral observations? How many profiles are optimal for environments with varying numbers of latent contexts? How should the assignment matrix $\boldsymbol{Z}$ adapt when the number or nature of contexts changes over time? These questions require systematic investigations of profile learnability under realistic data constraints and computational budgets.

\paragraph{Parameter estimation methodology.} We employed exhaustive grid search over discretized parameter spaces rather than continuous optimization methods such as maximum a posteriori (MAP) estimation or variational inference. This choice prioritized robustness—grid search systematically evaluates the entire parameter space and is insensitive to local minima or initialization effects—at substantial computational cost. Grid search scales poorly with the number of discretized parameters and profiles, growing rapidly as additional profile parameters or finer grids are introduced, which limits immediate extension to richer profile architectures. Gradient-based optimization methods would enable scaling to richer profile architectures but introduce challenges related to local optima and initialization sensitivity. The tradeoff between exhaustive search and computational efficiency represents a practical constraint on extending the framework to more complex scenarios.

\paragraph{Ecological validity and behavioral data.} We have not validated the framework against human or animal behavioral data. While our simulations demonstrate that the computational principle is sound and produces interpretable dynamics under latent context inference, establishing ecological validity requires fitting the model to actual choice sequences from reversal learning tasks. Do human subjects exhibit precision dynamics consistent with belief-weighted profile mixing? Can individual differences in profile recruitment predict measures of cognitive flexibility or clinical dimensions? Do patterns of hint-seeking behavior in human data reflect context-dependent strategic adaptation as predicted by the profile framework? Without empirical validation against real behavioral data, the framework remains a computational proof-of-concept rather than a validated model of biological decision-making. Integration with behavioral experiments, particularly tasks that manipulate reward probability structure or volatility regimes, would test whether the mechanisms we propose correspond to strategies employed by biological agents.

\paragraph{Model comparison scope.} Our comparison was limited to three active inference variants (M1, M2, M3) plus two non-Bayesian baselines. We did not compare against several relevant alternative families: (i) hierarchical active inference or volatility-inference models that maintain and update beliefs over higher-order environmental change parameters; (ii) Bayesian changepoint/oddball models that explicitly infer regime shifts via hazard-rate structure; or (iii) meta-learning and reinforcement-learning approaches that learn adaptation rules (e.g., learning-rate control) from experience. We focused on isolating the contribution of parameter sharing via belief-weighted recruitment of reusable value/control bundles; broader comparisons are needed to position profiles relative to these alternative adaptive mechanisms across task regimes.

\subsection{Future Directions}

The limitations outlined above suggest several promising directions for extending and validating the profile framework.

\paragraph{Complex inference scenarios and continuous contexts.} While we have demonstrated that profile mixing operates effectively when context must be inferred from discrete reward probability regimes, several extensions would test the framework's robustness under more challenging inference conditions. First, environments with continuous rather than discrete context variations—such as gradually changing volatility or reward rates—would require either continuous profile interpolation or dynamic profile generation. Can the framework be extended to allow smooth transitions across a continuous manifold of behavioral modes rather than discrete switching between fixed profiles? Second, scenarios with unknown or time-varying numbers of contexts would require online discovery and creation of new profiles. How should agents decide when to recruit existing profiles versus instantiating new ones? Third, integration with Bayesian changepoint detection or hierarchical inference over context transitions would enable the framework to handle environments where context boundaries are genuinely ambiguous and must be discovered through statistical inference rather than assumed a priori. Such extensions would test whether profile-based precision control scales beyond the relatively structured inference problems we have examined.

\paragraph{Profile learning and structure discovery.} Developing learning algorithms that discover profile number, parameter values, and state assignments from experience represents a critical next step toward practical applications. Variational approaches could treat profiles as latent variables with priors over their number and structure, enabling Bayesian model selection to determine optimal profile complexity. For instance, a Dirichlet process prior over profile assignments could allow the model to flexibly expand or contract the profile repertoire based on observed behavioral demands. Reinforcement learning methods could adapt profile parameters through reward-based feedback, allowing profiles to be tuned to environmental statistics rather than hand-designed. Testing whether such learning procedures converge to interpretable, reusable behavioral modes would validate that profile structure emerges naturally from experience rather than requiring experimenter specification. Critically, learned profiles should exhibit meaningful clustering in parameter space and interpretable associations with environmental statistics, providing evidence that the framework captures genuine structure rather than arbitrary parameterizations.

\paragraph{Hierarchical profile architectures.} Extending the single-level framework to hierarchical structures where meta-profiles govern subordinate profiles would capture nested goal structures and abstract contextual control. Higher-level profiles could modulate which lower-level profiles are accessible or preferred, creating compositional hierarchies of behavioral strategies. For example, a task-level profile might determine whether an agent operates in "foraging" versus "threat-avoidance" mode, while lower-level profiles within each mode specify context-appropriate action strategies. This would provide a more complete account of human goal-directed behavior while maintaining the computational benefits of parameterizing behavior through reusable modes. Critically, such hierarchical extensions would reconnect profile-based models with existing hierarchical active inference frameworks, demonstrating how profile mixing can be understood as an efficient parameterization of multi-level inference that reduces the number of free parameters required to capture adaptive behavior across multiple timescales and abstraction levels.

\paragraph{Behavioral and neural validation.} Fitting the framework to human behavioral data from reversal learning tasks would test whether profile recruitment patterns predict individual differences in cognitive flexibility, working memory capacity, or clinical dimensions such as compulsivity or exploration deficits. Tasks that manipulate reward probability structure (e.g., varying discrimination between contexts from subtle to obvious) could test whether human precision dynamics scale with the inferential challenge in ways predicted by belief-weighted profile mixing. Integration with neural and physiological measurements—including pupillometry as a proxy for noradrenergic tone, dopaminergic imaging, or thalamic activity patterns—could test whether computational mode switches align with known neural substrates of precision control and gain modulation. Such investigations would establish whether profile-based mechanisms reflect biological computation or merely provide a useful computational abstraction. Of particular interest would be testing the dissociation between preference-based and precision-based adaptation. Our results suggest that in certain volatile environments, strategic flexibility (modulating hint-seeking preferences) may be more effective than gain modulation (altering policy precision). Future behavioral experiments could test this prediction by distinguishing whether human subjects adapt to volatility by becoming "less confident" (stochastic action selection, consistent with precision modulation) or "more inquisitive" (directed information seeking, consistent with preference modulation). Differentiating these mechanisms would refine our understanding of how the brain resolves the exploration-exploitation dilemma.

\paragraph{Alternative parameter bundling structures.} The assumption that outcome preferences ($\boldsymbol{C}$), policy priors ($\boldsymbol{\xi}$), and policy precision ($\gamma$) should be bundled together reflects our hypothesis about which parameters co-vary during behavioral mode switching, but alternative groupings are theoretically possible. Systematic comparisons of models with different bundling structures—such as separating precision from preferences, including transition dynamics ($\boldsymbol{B}$) in profiles to capture different beliefs about environmental stability, or allowing state priors ($\boldsymbol{D}$) to vary across profiles—could determine which parameter combinations provide the most compact and interpretable account of adaptive behavior. Such investigations would refine our understanding of what constitutes a "behavioral mode" at the computational level and whether certain parameter groupings exhibit stronger empirical support or theoretical coherence than others.

\paragraph{Integration with context-dependent transition dynamics.} Our current implementation utilized a static transition matrix ($\boldsymbol{B}$) with fixed volatility assumptions to isolate the effects of value parameter adaptation. However, biological agents likely adapt both their value functions (what they want) and their world models (how the world changes) simultaneously. Future work should investigate "Regime Switching Active Inference," where profiles bundle transition dynamics ($\boldsymbol{B}_k$) alongside value parameters. For instance, a "Volatile Profile" could encode high transition uncertainty (flat $\boldsymbol{B}$ matrix), while a "Stable Profile" encodes deterministic transitions (sharp $\boldsymbol{B}$ matrix). This would allow agents to infer context from both reward statistics (as in our current study) and transition volatility, potentially improving adaptation speed in environments with subtle reward discrimination but distinct temporal dynamics.

\paragraph{Applications to computational psychiatry.} Profile-based models could provide computational phenotypes for psychiatric conditions characterized by atypical precision control or rigid behavioral strategies. Do patients with autism spectrum disorders exhibit reduced profile differentiation (using similar precision across contexts) or slower context-dependent recruitment (delayed updating of profile weights following context changes)? Do individuals with schizophrenia show excessive profile switching reflecting labile context beliefs or impaired ability to maintain stable behavioral modes? Do obsessive-compulsive symptoms correlate with overly rigid profile assignments that resist updating despite evidence for context changes? Investigating whether atypical profile dynamics correlate with clinical symptoms could connect computational mechanisms to psychiatric dimensions while suggesting intervention targets focused on improving adaptive strategy selection. Moreover, fitting the model to patient data could reveal whether clinical populations exhibit specific parameter alterations (e.g., flattened profile differentiation in autism, excessive profile uncertainty in schizophrenia) that produce testable predictions about behavioral performance and neural correlates.
\section{Conclusion}

A recurring challenge in predictive-processing and active-inference accounts is how to modulate value/control settings flexibly across latent contexts without maintaining independent parameters for every situation. This work introduced \emph{value profiles}---reusable bundles of outcome preferences, policy priors, and policy precision assigned to hidden states and recruited via belief-weighted mixing---as a computationally tractable parameter-sharing construction. In controlled model-recovery experiments, profile-based models were discriminable from global and entropy-coupled precision mechanisms, achieving substantially better penalized fit on profile-generated data (e.g., $\Delta$AIC $\approx 89$--108 under our setup). Parameter recovery further showed stable recoverability of the key structural feature within the searched parameterization even when context had to be inferred from noisy reward observations. Critically, within this task and model class, model-based mechanism attribution indicated that adaptation is expressed primarily through policy-prior differentiation ($\xi_{\text{hint}}: 6.0$ vs.\ $1.0$) rather than through context-specific changes in policy precision (constant $\gamma = 5.0$ in recovered fits), illustrating that profile-based control can recruit different computational channels depending on task structure.

The framework occupies a distinct position between static global control and fully hierarchical active inference: more structured than schemes that tie adaptation only to local uncertainty (e.g., entropy-coupled precision), yet more tractable than hierarchical architectures that introduce separate parameters across multiple timescales. By maintaining a small repertoire of behavioral modes and recruiting them through belief-weighted mixing over latent contexts---contexts that must be inferred from reward patterns rather than directly observed---agents can express context-appropriate strategy shifts with a compact set of reusable parameters. More broadly, the profile construction suggests a testable hypothesis: mode transitions should be expressed as systematic changes in action-type biases (policy priors/preferences) conditioned on inferred context identity, rather than solely as uniform changes in action stochasticity or decisiveness.

Overall, value profiles provide a concrete foundation for making state-conditional value control both compact and behaviorally recoverable, and they motivate future work on learning profile libraries and assignments from data, extending profiles to richer inference settings, and integrating profile-based value control with multi-level active-inference architectures.

\bibliographystyle{plainnat} 
\bibliography{references}

\end{document}